\documentclass{article}

\usepackage{arxiv}

\usepackage[utf8]{inputenc} 
\usepackage[T1]{fontenc}    
\usepackage{hyperref}       
\usepackage{url}            
\usepackage{booktabs}       
\usepackage{amsfonts}       
\usepackage{nicefrac}       
\usepackage{microtype}      
\usepackage{lipsum}		
\usepackage{graphicx}
\usepackage{natbib}
\usepackage{doi}
\usepackage{amsmath}

\usepackage{caption}
\usepackage{subcaption}

\title{Activation Functions: Dive into an optimal activation function }

\author{ \href{https://orcid.org/
0000-0003-1663-2670}{\includegraphics[scale=0.06]{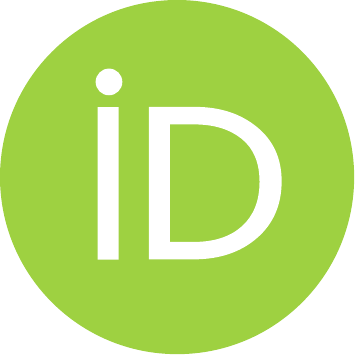}\hspace{1mm}Vipul Bansal}\\
	Department of
Mechanical and Industrial Engineering\\
	Indian Institute Of Technology–Roorkee\\
	Roorkee, India \\
	\texttt{vbansal@me.iitr.ac.in} \\
}

\renewcommand{\shorttitle}{Activation Functions}

\begin{document}
\maketitle

\begin{abstract}
	Activation functions have come up as one of the essential components of neural networks. The choice of adequate activation function can impact the accuracy of these methods. In this study, we experiment for finding an optimal activation function by defining it as a weighted sum of existing activation functions and then further optimizing these weights while training the network. The study uses three activation functions, ReLU, tanh, and sin, over three popular image datasets, MNIST, FashionMNIST, and KMNIST. We observe that the ReLU activation function can easily overlook other activation functions. Also, we see that initial layers prefer to have ReLU or LeakyReLU type of activation functions, but deeper layers tend to prefer more convergent activation functions. 
\end{abstract}

\keywords{Activation Functions \and Deep Learning \and Neural Networks }

\section{Introduction}
Deep Learning is currently one of the most booming fields(\cite{alzubaidi2021review}). Deep Neural Networks have been a standard component for all sorts of research done in the domain. These methods have acquired a wide variety of success in transforming the modern world. The development of many standard architectures has shown the effectiveness of these methods for various classification(\cite{SHARMA2018377}) and segmentation(\cite{segment}) tasks. 

Many models namely, AlexNet(\cite{Krizhevsky14}), VGG(\cite{simonyan2014very}), ResNet(\cite{he2016deep}), SqueezeNet(\cite{iandola2016squeezenet}), DenseNet(\cite{huang2018densely}), Inception (\cite{szegedy2015rethinking}), GoogLeNet(\cite{szegedy2014going}), etc., have gained popularity in recent years. This study focuses on one of the critical components in all the standard architectures, i.e., Activation Functions. Before going deep into the activation function, we need to understand how a neural network works. 

Lets consider a input $x_i$ such that we need to predict $y$ and estimate a function $f(x_i)$ with some error $\epsilon$ such that:
\begin{equation}
    y = f(x_i)+ \epsilon
\end{equation}
The simplest way we can find predicted value $\hat{y}$ is by writing $\hat{y} = w_i*x_i + b$, where $w_i$ stands for weight and $b$ stands for bais. now lets consider $X$ be a set of input features $[x_1 ~x_2 .....~ x_n]^T$ of shape $n\times1$. In this scenario $\hat{y}$ can be written as: 
\begin{equation}
    \hat{y}= [w_1~w_2~ w_3 .... ~w_n][x_1 ~x_2 .....~ x_n]^T + [b]
\end{equation}
Now, lets consider $y$ to be such that $[y_1 ~y_2 .....~ y_m]^T$ of shape $m\times1$. In such a case weight matrix $W_{(1)}$ has to be defined of shape $n\times m$ and bias $B_{(1)}$ is defined of shape $m\times1$. Therefore we can write the regression equation as: 
\begin{equation}
    \hat{y}= W_{(1)}^T  X +B_{(1)}
\end{equation}
Now, to model at a higher level we can further multiply another matrix $W_{(2)}$ with shape $n1\times m$ and bias $B_{(2)}$ of shape $m \times 1$. Also we modify  weight matrix $W_{(1)}$ has to be defined of shape $n\times n1$ and bias $B_{(1)}$ is defined of shape $n1\times1$. Hence, we can find $\hat{y}$ as:

\begin{equation}
    \hat{y}= W_{(2)}^T (W_{(1)}^T  X +B_{(1)})+B_{(2)}
\end{equation}
But just using additional matrix $W_{(2)}$, would not solve the problem of modelling difficult functions. Hence there is a need of a function that can add non-linearity to the model. Hence activation function $\mathcal{A}_{1}(.)$ and $\mathcal{A}_{2}(.)$ are required. Hence we can write final equation \ref{eq:network} as:
\begin{equation}
\label{eq:network}
    \hat{y}= \mathcal{A}_{2}(W_{(2)}^T \mathcal{A}_{1}(W_{(1)}^T  X +B_{(1)})+B_{(2)})
\end{equation}
The choice of activation function can actually be sometime quite tricky. There are wide variety of popular activation function namely, ReLU, Sigmoid, Tanh, GeLU, GLU(\cite{hendrycks2020gaussian}), Swish(\cite{ramachandran2017searching}), Softplus, Mish (\cite{misra2020mish}), PRelU(\cite{he2015delving}), SeLU(\cite{klambauer2017selfnormalizing}), Maxout(\cite{goodfellow2013maxout}), ReLU6(\cite{howard2017mobilenets}), HardSwish (\cite{howard2019searching}), ELU(\cite{clevert2016fast}), Shifted Softplus(\cite{schütt2017schnet}), SiLU(\cite{elfwing2017sigmoidweighted}), CReLU(\cite{shang2016understanding}), modReLU(\cite{arjovsky2016unitary}), KAF(\cite{scardapane2017kafnets}), TanhExp(\cite{liu2020tanhexp}) etc. 

\section{Experimental Setup}

In this work we are looking for a possibility of a better activation function by expressing it as a weighted sum of multiple activation functions and then optimizing those weights for them to find the dominant activation function. 

To start with we take 3 common machine learning datasets, MNIST(\cite{726791}), Fashion MNIST(\cite{xiao2017fashionmnist}) and KMNIST(\cite{abs-1812-01718}).  Further, we need to choose a network for experimentation.    The model used consist of two convolutional layers followed by two fully connected layers. There are activation functions between the layers which is weighted in format. We take 3 functions as activation function, ReLU($f_{1}$), Tanh($f_{2}$) and sin($f_{3}$) as given below:

\begin{equation}
    f_1(x)= ReLU(x)=max(o,x)
\end{equation}

\begin{equation}
    f_2(x)= tanh(x)=\frac{exp(x)-exp(-x)}{exp(x)+exp(-x)}
\end{equation}

\begin{equation}
    f_3(x)= Sin(x)
\end{equation}

Now further we define weights $w_1$, $w_2$, and $w_3$ such that  $w_1,w_2,w_3 \geq 0$. Using these we find parameters as $P_1$, $P_2$ and $P_3$ such that:

\begin{equation}
    P_i=\frac{w_i}{\Sigma w_i}, ~ for ~ i=1,2,3
\end{equation}
Finally we define activation function $A(.)$ as:
\begin{equation}
    A(x) = P_1 f_1(x) + P_2 f_2(x) + P_3 f_3(x)
\end{equation}
In the above equation $0\leq P_i\leq 1$ always. For each layer we define a set of unique parameters  $w_1,w_2,w_3$ for each activation function.

For training this network we adopt a unique strategy using Adam optimizer and adoption of three cycle of training. For the first cycle we start with freezing the weights $w_1,w_2,w_3$ and training the parameters of all other layers with a learning rate of $1e-3$ for 10 epochs. Then further freezing the parameters of convolutional and fully connected layers and allowing training of only $w_1,w_2,w_3$  with a learning rate of $1e-2$ for 10 epochs. At last,  we again  freeze the weights $w_1,w_2,w_3$ and train the parameters of all other layers with a learning rate of $1e-3$ for 10 epochs. We look into the final activation functions and weights for analysis.

\section{Results and Discussion}
\begin{figure}[h!]
     \centering
     \begin{subfigure}[b]{0.3\textwidth}
         \centering
         \includegraphics[width=\textwidth]{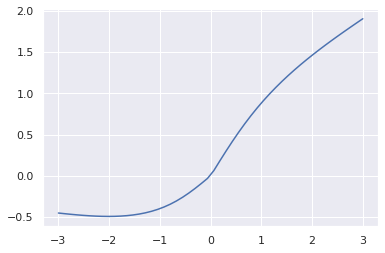}
         \caption{}
         \label{fig:y equals x}
     \end{subfigure}
     \hfill
     \begin{subfigure}[b]{0.3\textwidth}
         \centering
         \includegraphics[width=\textwidth]{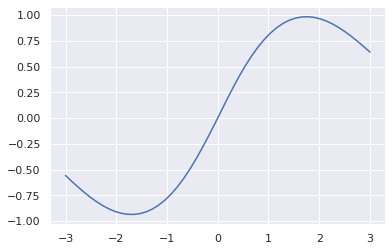}
         \caption{}
         \label{fig:three sin x}
     \end{subfigure}
     \hfill
     \begin{subfigure}[b]{0.3\textwidth}
         \centering
         \includegraphics[width=\textwidth]{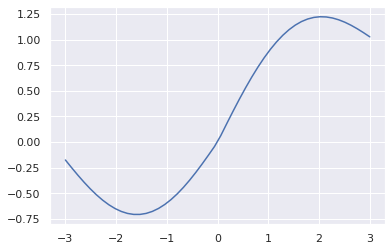}
         \caption{}
         \label{fig:five over x}
     \end{subfigure}

          \centering
     \begin{subfigure}[b]{0.3\textwidth}
         \centering
         \includegraphics[width=\textwidth]{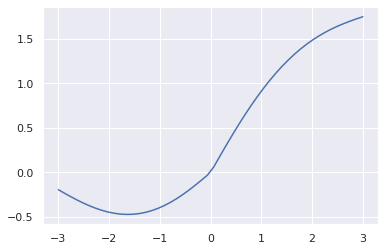}
         \caption{}
         \label{fig:y equals x}
     \end{subfigure}
     \hfill
     \begin{subfigure}[b]{0.3\textwidth}
         \centering
         \includegraphics[width=\textwidth]{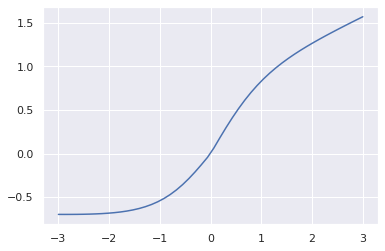}
         \caption{}
         \label{fig:three sin x}
     \end{subfigure}
     \hfill
     \begin{subfigure}[b]{0.3\textwidth}
         \centering
         \includegraphics[width=\textwidth]{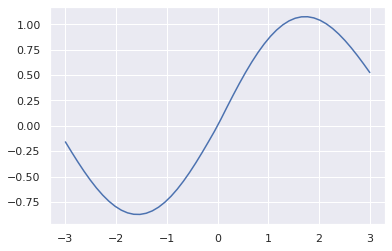}
         \caption{}
         \label{fig:five over x}
     \end{subfigure}

          \centering
     \begin{subfigure}[b]{0.3\textwidth}
         \centering
         \includegraphics[width=\textwidth]{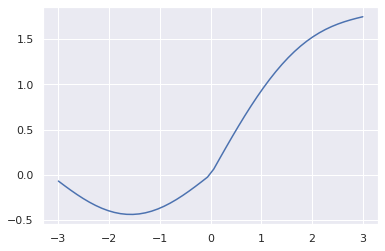}
         \caption{}
         \label{fig:y equals x}
     \end{subfigure}
     \hfill
     \begin{subfigure}[b]{0.3\textwidth}
         \centering
         \includegraphics[width=\textwidth]{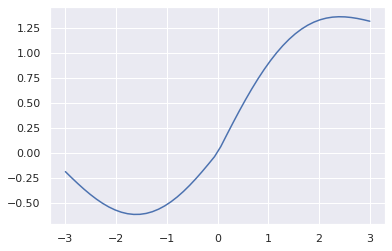}
         \caption{}
         \label{fig:three sin x}
     \end{subfigure}
     \hfill
     \begin{subfigure}[b]{0.3\textwidth}
         \centering
         \includegraphics[width=\textwidth]{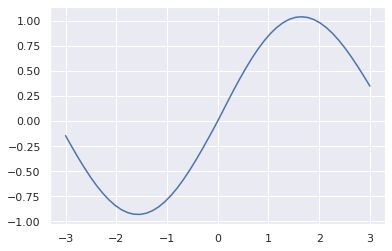}
         \caption{}
         \label{fig:five over x}
     \end{subfigure}
        \caption{Activation Function for MNIST((a) Layer 1, (b) Layer 2, (c) Layer 3)), Fashion MNIST ((d) Layer 1, (e) Layer 2, (f) Layer 3))and KMNIST((g) Layer 1, (h) Layer 2, (i) Layer 3)) }
        \label{fig:layers}
\end{figure}

In this section we discuss the results from various experiments on the three datasets. Table \ref{tab:result} shows the results obtained for all three datasets for all 3 layers of activation function for parameters $P_1$, $P_2$, and $P_3$. The Figure \ref{fig:layers} shows the activation functions for the corresponding layers for a range of value given $-3\leq x \leq 3$.  

We refer $A_1$ to be activation function of first layer of the network and write the same as for MNIST, Fashion MNIST and KMNIST:

\begin{equation}
\label{eq:MNIST-A1}
    A_1 = 0.4848 f_1 + 0.4437 f_2 + 0.0715 f_3~ for ~ MNIST
\end{equation}

\begin{equation}
\label{eq:fMNIST-A1}
    A_1 = 0.5178 f_1 + 0.1470 f_2 + 0.3352 f_3~ for ~ FashionMNIST
\end{equation}

\begin{equation}
\label{eq:kMNIST-A1}
    A_1 = 0.5590 f_1 + 0.0101 f_2 + 0.4309 f_3~ for ~ KMNIST
\end{equation}

\begin{table}[h!]
\caption{Weights $P_1, P_2, P_3$ for various activation layters for various datasets}
\vspace{1mm}
\centering
\label{tab:result}
\begin{tabular}{lllll}
\hline\\[1mm]
{DATASETS} & LAYERS & $P_1$ & $P_2$ & $P_3$ \\[1mm] \hline\\
 & LAYER 1 & 0.4848 & 0.4437 & 0.0715 \\
MNIST & LAYER 2 & 0.0276 & 0.4923 & 0.48 \\
 & LAYER 3 & 0.284 & 0.0877 & 0.6283 \\[1mm] \hline\\
 & LAYER 1 & 0.5178 & 0.147 & 0.3352 \\
FashionMNIST & LAYER 2 & 0.2907 & 0.7001 & 0.0091 \\
 & LAYER 3 & 0.1221 & 0.041 & 0.8369 \\[1mm] \hline\\
 & LAYER 1 & 0.559 & 0.0101 & 0.4309 \\
KMNIST & LAYER 2 & 0.3754 & 0.1167 & 0.5079 \\
 & LAYER 3 & 0.0679 & 0.0156 & 0.9165 \\[1mm] \hline
\end{tabular}
\end{table}

\begin{figure}[h!]
     \centering
     \begin{subfigure}[b]{0.3\textwidth}
         \centering
         \includegraphics[width=\textwidth]{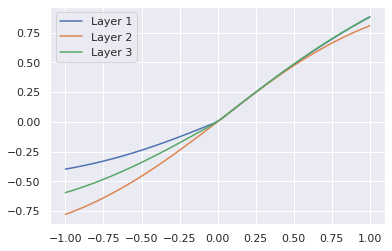}
         \caption{}
         \label{fig:y equals x}
     \end{subfigure}
     \hfill
     \begin{subfigure}[b]{0.3\textwidth}
         \centering
         \includegraphics[width=\textwidth]{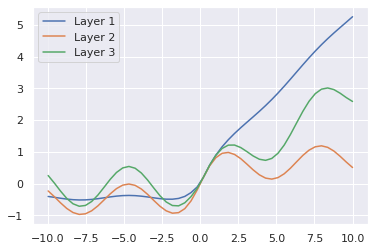}
         \caption{}
         \label{fig:three sin x}
     \end{subfigure}
     \hfill
     \begin{subfigure}[b]{0.3\textwidth}
         \centering
         \includegraphics[width=\textwidth]{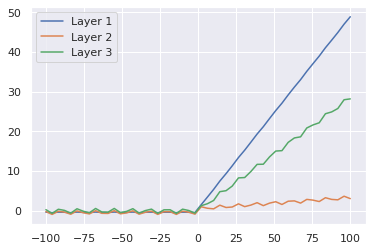}
         \caption{}
         \label{fig:five over x}
     \end{subfigure}

      \centering
     \begin{subfigure}[b]{0.3\textwidth}
         \centering
         \includegraphics[width=\textwidth]{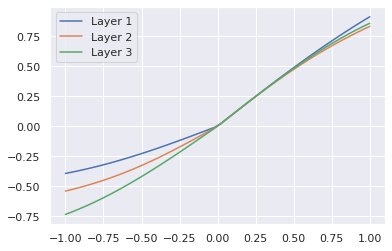}
         \caption{}
         \label{fig:y equals x}
     \end{subfigure}
     \hfill
     \begin{subfigure}[b]{0.3\textwidth}
         \centering
         \includegraphics[width=\textwidth]{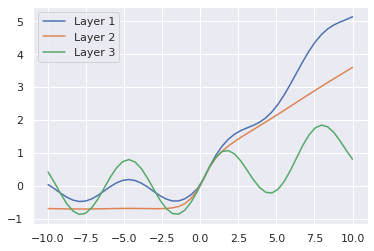}
         \caption{}
         \label{fig:three sin x}
     \end{subfigure}
     \hfill
     \begin{subfigure}[b]{0.3\textwidth}
         \centering
         \includegraphics[width=\textwidth]{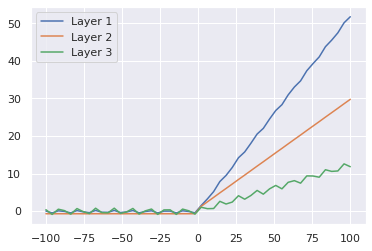}
         \caption{}
         \label{fig:five over x}
     \end{subfigure}

     \centering
     \begin{subfigure}[b]{0.3\textwidth}
         \centering
         \includegraphics[width=\textwidth]{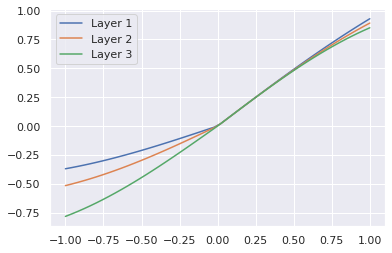}
         \caption{}
         \label{fig:y equals x}
     \end{subfigure}
     \hfill
     \begin{subfigure}[b]{0.3\textwidth}
         \centering
         \includegraphics[width=\textwidth]{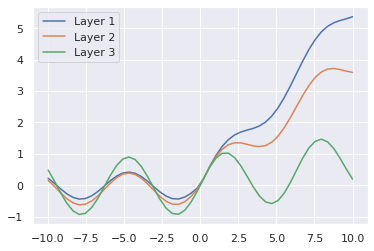}
         \caption{}
         \label{fig:three sin x}
     \end{subfigure}
     \hfill
     \begin{subfigure}[b]{0.3\textwidth}
         \centering
         \includegraphics[width=\textwidth]{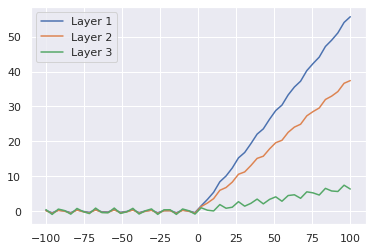}
         \caption{}
         \label{fig:five over x}
     \end{subfigure}
     
        \caption{Activation Function for MNIST((a) $-1 \leq x \leq 1$, (b) $-10 \leq x \leq 10$, (c)$-100 \leq x \leq 100$)), Fashion MNIST ((a) $-1 \leq x \leq 1$, (b) $-10 \leq x \leq 10$, (c)$-100 \leq x \leq 100$))and KMNIST((a) $-1 \leq x \leq 1$, (b) $-10 \leq x \leq 10$, (c)$-100 \leq x \leq 100$)) }
        \label{fig:range}
\end{figure}

From the equation \ref{eq:MNIST-A1}, \ref{eq:fMNIST-A1} and \ref{eq:kMNIST-A1} we can observe that first activation layer is more dominated by  ReLU ($f_1$) activation function. This shows that for first layer of a network it would always be prefered to choose an activation function similar to ReLU activation function.   When we further take a look into activation layers $A_2$ and $A_3$, we observe that the cofficient of ReLU activation function is smaller than that of other activation functions.

\begin{equation}
\label{eq:MNIST-A2}
    A_2 = 0.0276 f_1 + 0.4923 f_2 + 0.4800 f_3~ for ~ MNIST
\end{equation}

\begin{equation}
\label{eq:fMNIST-A2}
    A_2 = 0.2907 f_1 + 0.7001 f_2 + 0.0091 f_3~ for ~ FashionMNIST
\end{equation}

\begin{equation}
\label{eq:kMNIST-A2}
    A_2 = 0.3754 f_1 + 0.1167 f_2 + 0.5079 f_3~ for ~ KMNIST
\end{equation}

\begin{equation}
\label{eq:MNIST-A3}
    A_3 = 0.2840 f_1 + 0.0877 f_2 + 0.6283 f_3~ for ~ MNIST
\end{equation}

\begin{equation}
\label{eq:fMNIST-A3}
    A_3 = 0.1221 f_1 + 0.0410 f_2 + 0.8369 f_3~ for ~ FashionMNIST
\end{equation}

\begin{equation}
\label{eq:kMNIST-A3}
    A_3 = 0.0679 f_1 + 0.0156 f_2 + 0.9165 f_3~ for ~ KMNIST
\end{equation}

After we observe the activation function for individual layer we look for observing them in various ranges of input. When we take $-1 \leq x \leq 1$ than we observe the plot in figure \ref{fig:range} (a), (d), and (e). We find that activation function can be approximately written as:

\begin{equation}
    A=
    \begin{cases}
      h_1*x, & if ~x\geq 0  \\
       h_2*x, & \text{otherwise}
    \end{cases}
\end{equation}
In the above activation function $h_1$ and $h_2$ are arbitrary small constants. The above function can be transfered and written as LeakyReLU activation function by taking $h_1=1$ and $h_2$ be a small value. hence it can be written as: 
\begin{equation}
    A= max(h_2*x,x) = LeakyReLU(X)
\end{equation}

If we take a broader look these approximations are good for activation function in initial layer of the network, but as we go deeper we observe that more convergent activation functions start becoming dominant. But, when $x$ starts to be larger we observe that ReLU can easily dominate other activation functions because of it larger value as shown in Figure \ref{fig:range} (c), (f), and (i).

\section{Conclusion}
This work shows an experimental study of utilizing a weighted linear combination of activation functions, looking for a better activation function needed by the model. We observe that the ReLU activation function is a powerful activation function that can overlook other activation functions when used in conjunction. We also follow that initial activation layers prefer a ReLU type or LeakyReLU activation function. We observe that networks prefer more convergent activation functions like tanh, sigmoid or sin for deeper layers. We can further explore the possibility of utilizing more activation functions and experimenting with more datasets. Additionally, it is possible to use multiplicative combinations of activation functions that would help us find better and more robust activation functions for neural networks.

\bibliographystyle{unsrtnat}
\bibliography{references}

\end{document}